\documentclass[]{fairmeta}

\usepackage{amsmath}
\usepackage{amssymb}
\usepackage{mathtools}
\usepackage{amsthm}
\usepackage{comment}
\usepackage{siunitx}
\usepackage{pifont}
\usepackage{adjustbox}
\usepackage{makecell}
\usepackage{wrapfig}
\usepackage{algorithm}
\usepackage{algorithmicx}
\usepackage{algpseudocode}
\usepackage{listings}
\usepackage{manyfoot}
\usepackage[title]{appendix}
\usepackage{mathrsfs}
\usepackage{booktabs} 
\usepackage{tabularx} 
\usepackage[table]{xcolor} 
\usepackage{colortbl}

\theoremstyle{plain}

\theoremstyle{definition}

\theoremstyle{remark}

\title{%
  \raisebox{-0.15em}{\includegraphics[height=1.5em]{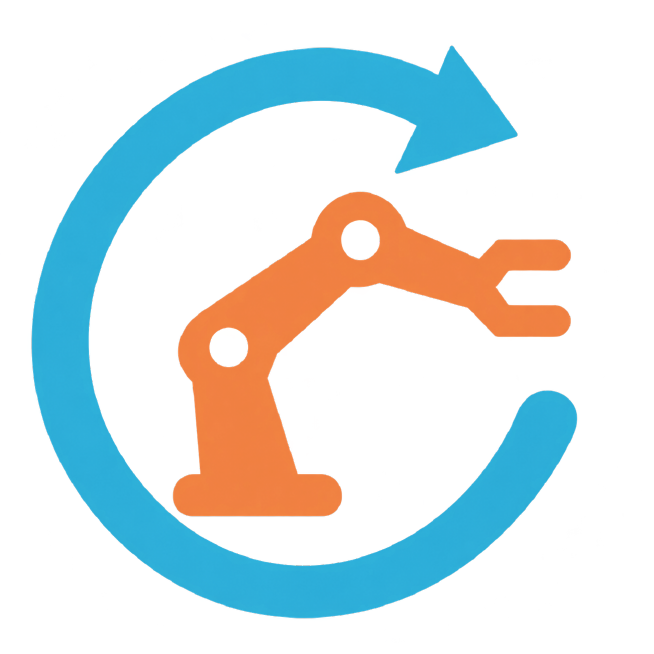}}%
  \hspace{0.2em}Embodied-Harness: Towards Recursive Self-\\
  Improvement for Robotic Manipulation%
}

\author[1]{Zhenxuan Fan}
\author[2]{Zhuoya Wang}
\author[1]{Yuqian Yuan}
\author[1\dagger]{Wenqiao Zhang}
\author[1]{Juncheng Li}
\author[1]{Jun Xiao}
\author[1]{Siliang Tang}
\author[1]{Yueting Zhuang}

\affiliation[1]{Zhejiang University}
\affiliation[2]{Huazhong University of Science and Technology}
\contribution[\dagger]{Corresponding author}

\abstract{
Vision-language model (VLM)-based agents offer
a promising approach to robotic manipulation,
yet remain limited by inadequate memory organization,
uncertain failure attribution, and limited
reuse of execution experience.
We propose \textbf{Embodied-Harness}, a
dual-agent framework that improves manipulation
by refining the execution harness without updating
model parameters.
The Actor interacts with the environment through
a robot-control interface, while the Reflector
converts execution evidence into reusable
operational guidance.
Hierarchical Memory Architecture maintains short-term execution
context and long-term experiential knowledge
to support execution and reflection.
Evidence-based Experience Reflection grounds failure diagnosis
in observed interactions, while Recursive Harness
Self-Improvement refines operational guidance
across successive attempts.
Experiments on RoboTwin 2.0, RoboSPA, and RMBench
demonstrate the effectiveness of Embodied-Harness
in generalization under environmental variations,
spatial reasoning, long-horizon planning, and
memory-dependent manipulation.
Our method achieves success rates of \textbf{82.4\%},
\textbf{76.7\%}, and \textbf{94.4\%} on the three benchmarks,
respectively.
Real-world evaluation further yields an average
success rate of \textbf{96.7\%} across three manipulation tasks.
These results highlight the potential of
Embodied-Harness to improve robotic manipulation
through experience-driven harness self-improvement.}

\renewcommand{\titlefigure}{%
  \par\vspace{-0.3em}
  {\raggedright\small\noindent
  \raisebox{-0.15em}{\includegraphics[height=1em]{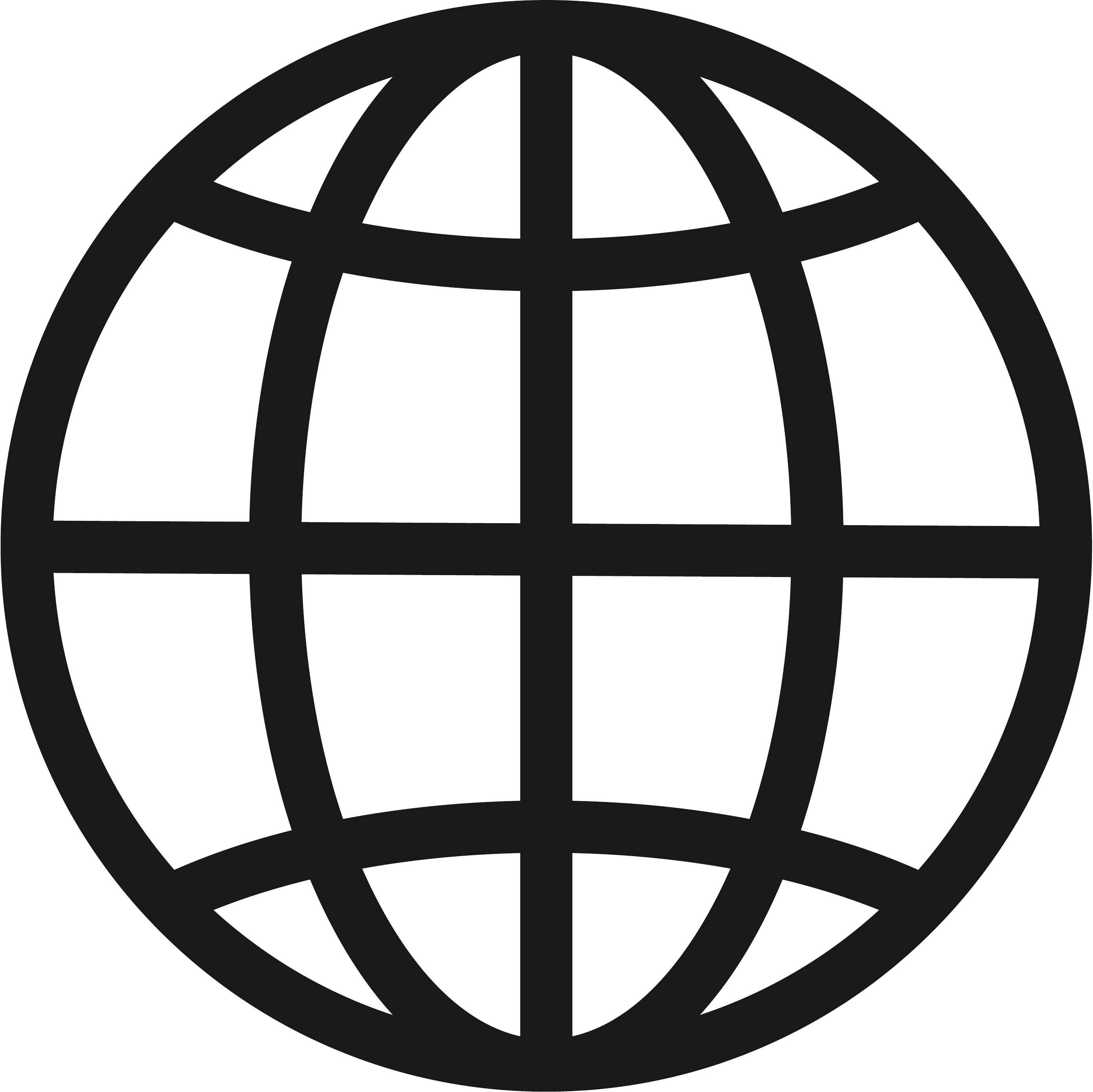}}%
  \hspace{0.3em}\textbf{Project website:}~\url{https://embodied-harness.github.io/}\par}
  \vspace{0.1em}
  \noindent\begin{minipage}{\linewidth}
    \centering
    \includegraphics[width=\linewidth]{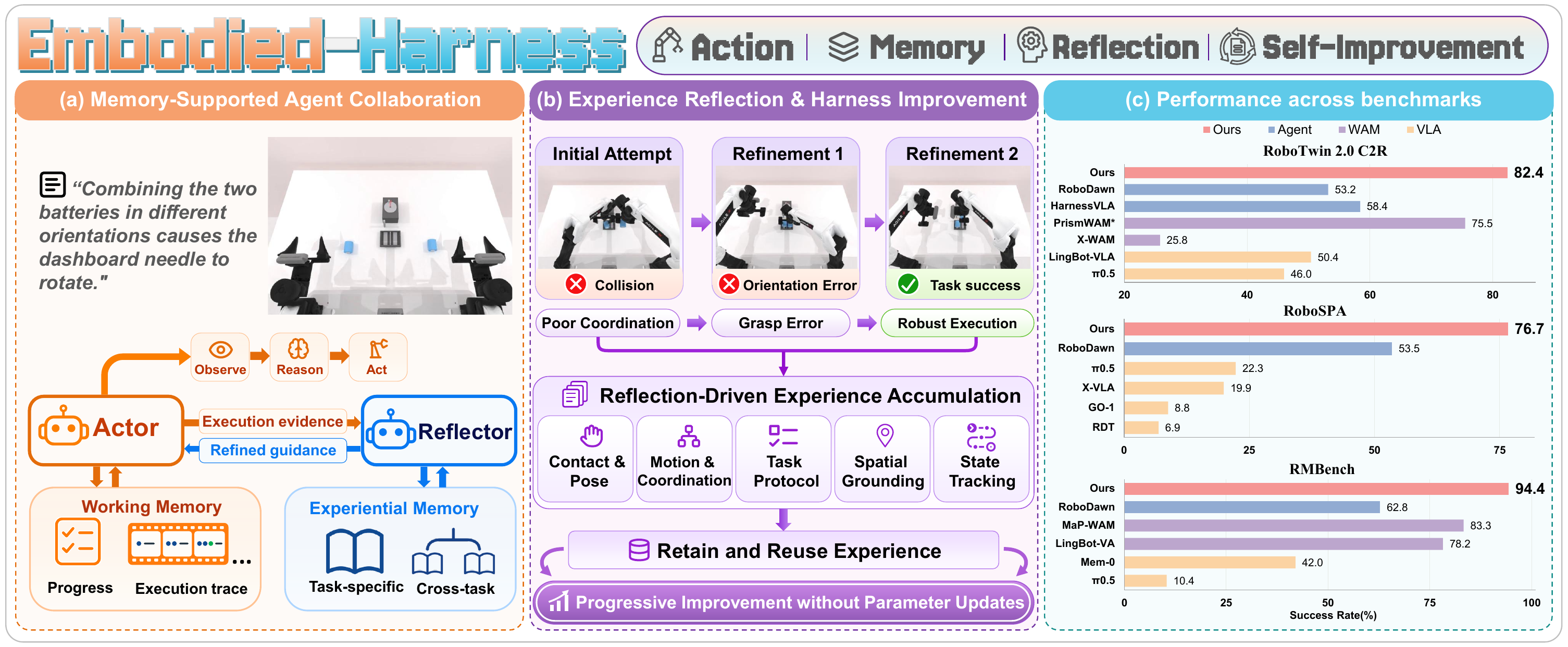}   
\captionof{figure}{\textbf{Overview of Embodied-Harness}.
\textbf{(a)} Memory-supported collaboration between
the Actor and Reflector for task execution,
experience analysis, and guidance generation.
\textbf{(b)} Experience-driven reflection and
recursive harness self-improvement, progressively
correcting execution failures through evidence-based
guidance without model parameter updates.
\textbf{(c)} Performance comparison on
RoboTwin 2.0, RoboSPA, and RMBench,
demonstrating the effectiveness of Embodied-Harness
across diverse manipulation benchmarks.
}
    \label{fig:intro}
  \end{minipage}%
  \par
}

\begin{document}


\maketitle

\section{Introduction}
\label{sec:intro}

Robotic manipulation is a fundamental capability of
embodied intelligence~\cite{doi:10.1126/science.aat8414,JMLR:v22:19-804,9687596}, enabling robots to interact
with the physical world and perform diverse tasks
across household assistance~\cite{pmlr-v205-li23a,pmlr-v270-fu25b}, industrial manufacturing~\cite{Narang-RSS-22,Tang-RSS-23},
and healthcare~\cite{doi:10.1126/scirobotics.abi8017,pmlr-v270-kim25a}.
Developing general manipulation capabilities
is essential for enabling versatile and autonomous
robotic operation in real-world environments,
representing a critical step toward general
embodied intelligence (GEI)~\cite{doi:10.1177/02783649241281508,11098567,geminiroboticsteam2025geminiroboticsbringingai}.

Towards this goal, different paradigms have emerged
for robotic manipulation.
End-to-end frameworks, including vision-language-action
(VLA) models~\cite{pmlr-v229-zitkovich23a,pmlr-v270-kim25c,nvidia2025gr00tn1openfoundation,intelligence2025pi05visionlanguageactionmodelopenworld,shukor2025smolvlavisionlanguageactionmodelaffordable,zheng2026xvla,wu2026pragmaticvlafoundationmodel} and world-action models (WAMs)~\cite{pmlr-v267-hu25g,Bi_2026_CVPR,ICLR2026_748becc4,LiL-RSS-26,ye2026worldactionmodelszeroshot,yuan2026fastwamworldactionmodels}, have
demonstrated remarkable progress in learning robotic
policies from multimodal inputs.
However, these approaches often rely on large-scale
robotic datasets, face challenges in generalizing
to unseen environments~\cite{10611477,Khazatsky-RSS-24,Ma_2026}, and struggle with reliable
long-horizon manipulation~\cite{9788026,NEURIPS2023_8c3c6668,fan-etal-2026-robospa}.
In contrast, agentic frameworks leverage the reasoning
and planning capabilities of vision-language models
(VLMs) to decompose complex tasks and coordinate
robot actions~\cite{pmlr-v205-ichter23a,10160591,pmlr-v229-huang23b,wang2026embodiedskillsworldactionmodelaugmented}.
Such frameworks support flexible action planning and
feedback-driven execution, offering a promising
alternative without requiring task-specific updates
to the underlying model parameters~\cite{pmlr-v205-huang23c,pmlr-v270-duan25a,zhang2026harnessvlasteeringfrozen}.

Despite these advantages, agentic frameworks still face
several challenges in achieving reliable robotic
manipulation.
\textbf{(i) Inadequate Memory Organization.}
Many real-world manipulation tasks are inherently
long-horizon, requiring agents to track task progress
and detailed execution histories while retaining
experience across attempts and tasks~\cite{10.1145/3586183.3606763,DBLP:journals/tmlr/WangX0MXZFA24,10.1145/3748302,chen2026rmbenchmemorydependentroboticmanipulation}.
Managing such information across different temporal
scales and levels of granularity remains challenging.
\textbf{(ii) Uncertain Failure Attribution.}
Manipulation failures may stem from inaccurate action
grounding, unsatisfied preconditions, or violated
state constraints~\cite{pmlr-v229-liu23g,ICLR2025_70a06501,Xia_2025_CVPR}.
However, execution feedback often reveals only
observable outcomes rather than underlying causes,
making it difficult to identify critical errors
and derive reliable operational guidance.
\textbf{(iii) Limited Self-Improvement Capability.}
Even when agents identify execution errors and
extract useful insights, these reflections may remain
isolated from subsequent executions.
Without mechanisms to accumulate, validate, and
refine operational guidance across attempts,
agents may repeatedly encounter similar failures~\cite{NEURIPS2023_1b44b878,Zhao_Huang_Xu_Lin_Liu_Huang_2024,NEURIPS2024_0142921f}.

To address these challenges, we propose
\textbf{Embodied-Harness}, a dual-agent framework
for robotic manipulation that enables recursive
harness self-improvement through accumulated
execution experience (Fig.~\ref{fig:intro}).
The framework comprises two complementary roles,
an \textbf{Actor} and a \textbf{Reflector}.
The Actor interacts with the environment through
control interfaces, while the Reflector analyzes
execution trajectories and environmental feedback
to improve subsequent task execution.
Specifically, Embodied-Harness incorporates three
complementary mechanisms:
\textbf{(i) Hierarchical Memory Architecture}
organizes short-term execution context and long-term
experiential knowledge across different temporal scales;
\textbf{(ii) Evidence-based Experience Reflection}
identifies potential failure causes from execution
evidence and derives actionable operational guidance;
and \textbf{(iii) Recursive Harness Self-Improvement}
refines the execution harness using accumulated
guidance, with subsequent execution feedback
informing further reflection and refinement.
Together, these mechanisms enable continual
improvement in robotic manipulation without updating
the underlying model parameters.


We evaluate Embodied-Harness on three robotic
manipulation benchmarks, RoboTwin 2.0, RoboSPA,
and RMBench, covering generalization under
environmental variations, fine-grained spatial
reasoning, long-horizon procedural planning, and
memory-dependent manipulation.
Embodied-Harness outperforms baselines spanning
VLA models, WAMs, and agentic frameworks,
achieving average success rates of \textbf{82.4\%},
\textbf{76.7\%}, and \textbf{94.4\%} on the three benchmarks,
respectively.
Real-world evaluation further yields an average
success rate of \textbf{96.7\%} across three manipulation
tasks.
Ablation studies and further analyses
support the effectiveness of the proposed
components and illustrate how accumulated
experience enables more reliable manipulation.

Our main contributions are summarized as follows:
\begin{itemize}
    

    \item We propose Embodied-Harness, an experience-driven
    dual-agent framework that improves robotic
    manipulation through complementary execution
    and reflection without updating model parameters.

    \item We design a Hierarchical Memory Architecture
    that integrates short-term working memory and
    long-term experiential memory to support coherent
    task execution and effective experience reuse.

    \item We introduce Evidence-based Experience
    Reflection and Recursive Harness Self-Improvement
    to derive actionable guidance from execution
    feedback and refine the harness across
    successive attempts.
    
    \item We conduct extensive experiments on three
    robotic manipulation benchmarks and real-world
    tasks, demonstrating the effectiveness of
    Embodied-Harness across diverse manipulation
    settings.
\end{itemize}

\section{Related Work}
\label{sec:related_work}

\subsection{End-to-End Models for Robotic Manipulation}
End-to-end policies map visual observations and language instructions directly to robot actions. Vision-language-action (VLA) models build this mapping on large-scale robot data and pretrained vision-language backbones~\cite{Brohan-RSS-23,pmlr-v229-zitkovich23a,Ghosh-RSS-24,pmlr-v270-kim25c}, with later works adopting diffusion or flow-matching action generators~\cite{liu2025rdtb,BlackK-RSS-25,nvidia2025gr00tn1openfoundation,intelligence2025pi05visionlanguageactionmodelopenworld,pan2026ironmindscalinghumanoiddexterous,bu2026movenszeroscalinghumanoidmotion} and extending them with cross-embodiment training, efficiency, or visual reasoning~\cite{agibotworldcontributors2025agibotworldcolosseolargescale,zheng2026xvla,shukor2025smolvlavisionlanguageactionmodelaffordable,wu2026pragmaticvlafoundationmodel,Zhao_2025_CVPR,wang2026jevvlajevinspiredenergybasedadaptation,gao2026visualthinkvlavisualintermediatereasoning}. World action models (WAMs) instead exploit video generative priors, either guiding actions with predicted futures~\cite{NEURIPS2023_1d5b9233,ICLR2024_2c37c5bc,pmlr-v267-hu25g} or jointly modeling video and actions~\cite{Bi_2026_CVPR,ICLR2026_748becc4,LiL-RSS-26,ye2026worldactionmodelszeroshot,yuan2026fastwamworldactionmodels,wang2026openwamopenmodularexploration}.
Despite their progress, these models typically
require large-scale robotic training data and
remain limited in generalizing to unseen
environments and task variations.

\subsection{Agentic Frameworks for Robotic Manipulation}
Agentic frameworks connect foundation models to robots through perception, planning, and control interfaces. Language models sequence skills, generate plans or programs, and incorporate environmental feedback~\cite{pmlr-v205-ichter23a,pmlr-v205-huang23c,10161317,zeng2023socratic,10160591}, while embodied VLMs predict plans, affordances, and spatial targets~\cite{pmlr-v202-driess23a,pmlr-v270-yuan25c,Ji_2025_CVPR,geminiroboticsteam2025geminiroboticsbringingai,dang2026rynnbrainopenembodiedfoundation}. Other methods translate VLM reasoning into value maps, keypoints, or geometric constraints for motion generation~\cite{pmlr-v229-huang23b,pmlr-v270-huang25g,Fang-RSS-24,10801352,Pan_2025_CVPR,pmlr-v270-duan25a}, or pair high-level reasoners with low-level policies~\cite{nvidia2025gr00tn1openfoundation,pmlr-v267-shi25d,gigabrainteam2026gigabrain07scalingembodiedfoundation,sensenova2026gpt6astraachievedembodiedintelligence,lin2026agentic,wang2026embodiedskillsworldactionmodelaugmented}. Closest to our setting, RoboDawn~\cite{guo2026transferringintelligencevlmsrobotic} controls robots through VLM-issued commands, while Harness VLA~\cite{zhang2026harnessvlasteeringfrozen} combines analytic primitives with a frozen VLA through memory-guided planning.
Embodied-Harness advances these frameworks through
coordinated execution and reflection, enabling
evidence-based failure diagnosis, experience reuse,
and progressive refinement of execution strategies
across attempts.

\subsection{Reflection and Self-Improvement in Agents}
Reflection lets agents improve from feedback without updating model parameters. Language agents reason while acting~\cite{yao2023react}, revise their own outputs~\cite{NEURIPS2023_91edff07}, or store verbal reflections across trials~\cite{NEURIPS2023_1b44b878}, while others distill experience into insights, memories, skills, or rules~\cite{Zhao_Huang_Xu_Lin_Liu_Huang_2024,10.1145/3586183.3606763,DBLP:journals/tmlr/WangX0MXZFA24,NEURIPS2024_0142921f,pmlr-v306-zhan26a}. Prompt and agent-design optimization further improves systems built on fixed models~\cite{ICLR2024_3339f19c,ICLR2024_f1cf02ce,Yuksekgonul2025,ICLR2025_36b7acf6,ICLR2025_5492ecbc}. In robotics, prior work explains, detects, or corrects manipulation failures~\cite{pmlr-v229-liu23g,ICLR2025_70a06501,Xia_2025_CVPR}, and SenseNova-RoboRSI~\cite{sensenova2026gpt6astraachievedembodiedintelligence} revises the agent harness and retains changes validated by re-evaluation. 
Embodied-Harness emphasizes the link between
failure diagnosis and corrective guidance,
using observed execution evidence to identify
potential failure causes and specify actionable
execution conditions for subsequent robotic
manipulation attempts.

\begin{figure*}[!t]
    \centering
 \includegraphics[width=\textwidth,height=0.40\textheight,keepaspectratio]{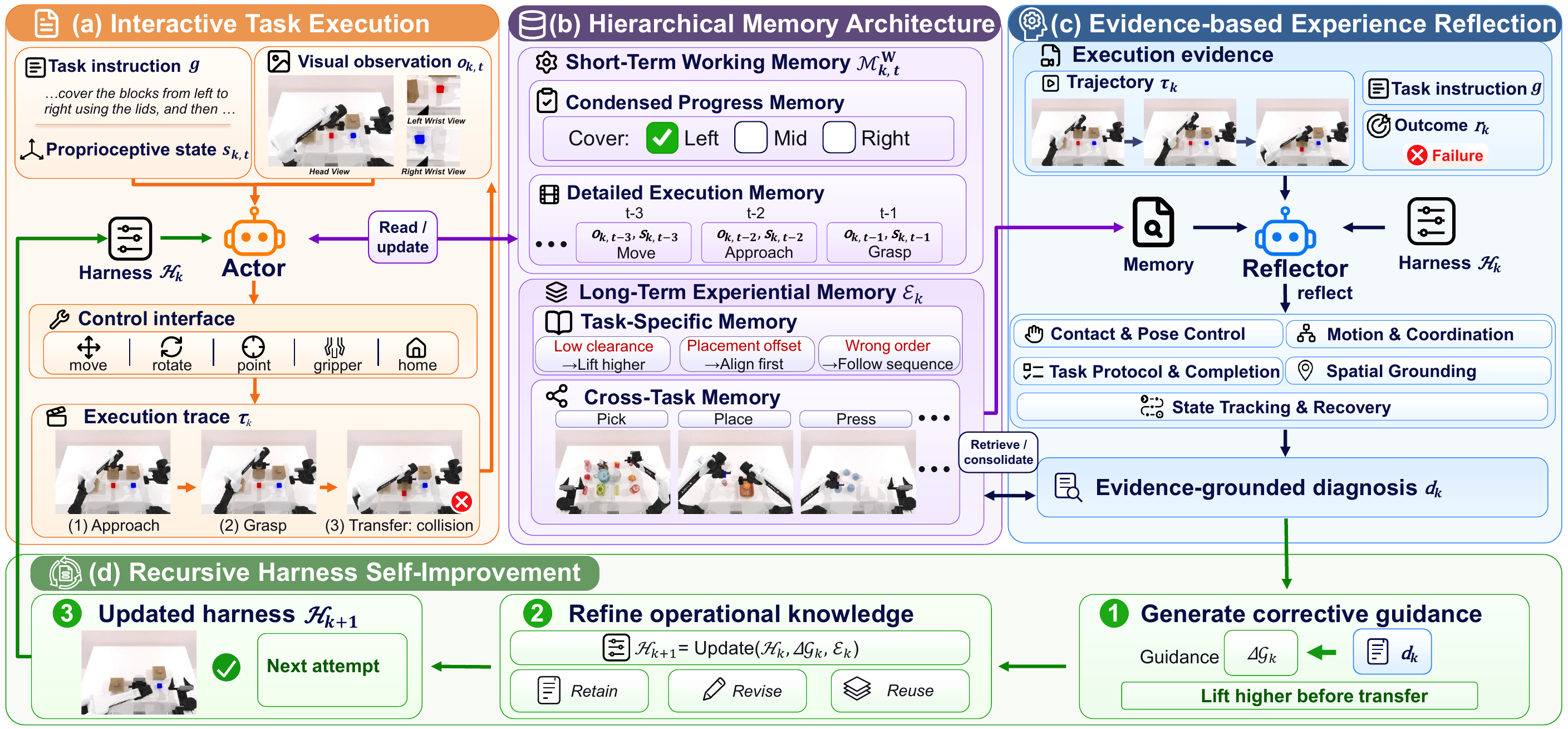} 
\caption{
\textbf{Framework of Embodied-Harness.}
\textbf{(a)} The Actor performs task
execution through a perception--action loop
guided by the execution harness.
\textbf{(b)} Hierarchical memory maintains
short-term execution context and long-term
experience across attempts and tasks.
\textbf{(c)} The Reflector analyzes execution
evidence and experience to
diagnose failures and identify improvements.
\textbf{(d)} Recursive harness self-improvement
generates guidance and refines
operational knowledge to improve subsequent
execution without updating model parameters.
}
    \label{fig:overview}
\end{figure*}


\section{Method}
\label{sec:method}

\subsection{Framework Overview}
\label{sec:framework}

We propose \textbf{Embodied-Harness}, a dual-agent
framework that improves robotic manipulation
through accumulated execution experience.
As illustrated in Fig.~\ref{fig:overview},
Embodied-Harness consists of two complementary
agents, an \textbf{Actor} and a \textbf{Reflector}, which
collaborate through a recursive
execution--reflection process.
The framework integrates Hierarchical Memory Architecture,
Evidence-based Experience Reflection, and Recursive Harness Self-Improvement to enable progressive improvement
without updating the underlying model parameters.


\noindent\textbf{Agent Role Specialization.}
Embodied-Harness separates task execution from
experience reflection.
The Actor interacts with the environment through
control interfaces to carry out the task, while
the Reflector examines execution outcomes to
improve subsequent attempts.
Drawing on execution evidence and accumulated
experience, the Reflector diagnoses potential
failure causes and generates corrective guidance
to refine the execution harness.
The updated harness then guides the Actor,
forming a closed loop of execution, reflection,
and self-improvement.

\noindent\textbf{Interactive Action Execution.}
Given a task instruction $g$, the Actor selects
a control command and grounded arguments based
on visual observations, proprioceptive states,
the current harness, and working memory:
\begin{equation}
    a_{k,t}\sim\pi_{\theta}\left(
    \cdot\mid g,o_{k,t},s_{k,t},
    \mathcal{H}_k,\mathcal{M}_{k,t}^{\mathrm{W}}
    \right),
\end{equation}
where $o_{k,t}$ and $s_{k,t}$ denote the visual
observation and proprioceptive state at step $t$
of execution attempt $k$, respectively.
Here, $a_{k,t}$ denotes the selected action,
$\pi_{\theta}$ is the frozen Actor model,
$\mathcal{H}_k$ denotes the current execution
harness, and $\mathcal{M}_{k,t}^{\mathrm{W}}$
represents the Actor's short-term working memory,
detailed in Sec.~\ref{sec:memory}.
The action interface supports five types
of control commands~\cite{guo2026transferringintelligencevlmsrobotic}:
\[
    \mathcal{A}_{\mathrm{type}}
    =
    \left\{
    \texttt{move},\;
    \texttt{rotate},\;
    \texttt{point},\;
    \texttt{gripper},\;
    \texttt{home}
    \right\}.
\]
These commands support end-effector translation,
rotation, predefined orientation control,
gripper actuation, and returning to the home pose,
respectively.

For commands involving end-effector motion,
the harness converts the action arguments into
a target pose $\hat{X}_{k,t}\in SE(3)$.
The corresponding inverse kinematics (IK)
problem is formulated as
\begin{equation}
    \mathbf{q}_{k,t}^{\star}
    \in
    \arg\min_{\mathbf{q}\in\mathcal{Q}_{\mathrm{valid}}}
    D\!\left(
        \mathrm{FK}(\mathbf{q}),
        \hat{X}_{k,t}
    \right),
\end{equation}
where $\mathrm{FK}$ denotes forward kinematics,
$D$ measures translational and rotational
pose discrepancies, and $\mathcal{Q}_{\mathrm{valid}}$
denotes the set of joint configurations satisfying
joint limits and collision constraints.
The resulting joint configuration is used
to plan a trajectory for robot execution.
After each action, the Actor uses feedback
$f_{k,t}$ and updated observations and states
to track progress and select subsequent actions.
The trajectory
$\tau_k=\{(o_{k,t},s_{k,t},a_{k,t},f_{k,t})\}_{t=1}^{T_k}$
provides evidence for reflection and harness refinement.

\subsection{Hierarchical Memory Architecture}
\label{sec:memory}

Robotic manipulation requires maintaining
execution context within individual episodes
while preserving reusable knowledge across
attempts and tasks. To support these different
temporal scales and information granularities,
we introduce a Hierarchical Memory Architecture
comprising Short-Term Working Memory and
Long-Term Experiential Memory.

\noindent\textbf{Short-Term Working Memory.}
Short-Term Working Memory supports the Actor's
decision-making within a single execution
episode by continuously maintaining task
progress and execution context. It comprises
\textit{Condensed Progress Memory} and
\textit{Detailed Execution Memory}, which
capture high-level task states and fine-grained
interaction histories, respectively.

\textit{Condensed Progress Memory.}
Long-horizon manipulation involves interdependent
subgoals whose progress is often scattered
across interaction histories, making it
difficult to maintain coherent task states.
To address this, \textit{Condensed Progress Memory}
consolidates completed subgoals, pending
operations, environmental changes, and
execution constraints into a compact
representation:
\begin{equation}
    \mathcal{M}_{k,t+1}^{c}
    = \operatorname{Summarize}
    \left(\mathcal{M}_{k,t}^{c},e_{k,t}\right),
\end{equation}
where $e_{k,t}=(o_{k,t},s_{k,t},a_{k,t},f_{k,t})$
denotes the interaction record at step $t$
of attempt $k$.
This enables the Actor to
track task progress without repeatedly
reconstructing it from historical interactions.

\textit{Detailed Execution Memory.}
Although condensed representations maintain
task progress, they may omit critical
action-level details needed to identify
and avoid previous mistakes.
\textit{Detailed Execution Memory} therefore preserves
chronological records of visual observations,
proprioceptive states, actions, and
execution feedback:
\begin{equation}
    \mathcal{M}_{k,t+1}^{d}
    = \mathcal{M}_{k,t}^{d}
    \oplus e_{k,t},
\end{equation}
where $\oplus$ denotes the append operation.
This memory supports feedback-driven action
adjustment and provides execution evidence
for the Reflector's failure analysis.

Together, these complementary components
constitute the Actor's short-term working memory:
\begin{equation}
    \mathcal{M}_{k,t}^{\mathrm{W}}
    =
    \left(
        \mathcal{M}_{k,t}^{c},
        \mathcal{M}_{k,t}^{d}
    \right).
\end{equation}

\noindent\textbf{Long-Term Experiential Memory.}
Long-Term Experiential Memory primarily serves
the Reflector by preserving experience across
episodes, providing historical context for
failure diagnosis and subsequent self-improvement.
It comprises \textit{Task-Specific Memory}
and \textit{Cross-Task Memory}, supporting
task-dependent refinement and transferable
knowledge reuse, respectively.

\textit{Task-Specific Memory.}
Repeated attempts at the same task may
encounter similar failures, motivating the
retention of previous execution experience.
\textit{Task-Specific Memory} records historical
outcomes, diagnosed errors, and corrective
guidance associated with each task:
\begin{equation}
    \mathcal{E}_{k+1}^{s}(i)
    = \operatorname{Update}\left(
    \mathcal{E}_{k}^{s}(i),
    r_k,d_k,\Delta\mathcal{G}_k
    \right),
\end{equation}
where $i$ denotes the task identity, shared
across episodes with potentially different
instructions, and $r_k$, $d_k$, and
$\Delta\mathcal{G}_k$ denote the task outcome,
execution diagnosis, and generated guidance,
respectively.
This memory helps identify recurring
failure patterns and supports targeted improvement across attempts.

\textit{Cross-Task Memory.}
Different manipulation tasks often share
transferable operational principles.
\textit{Cross-Task Memory} captures such knowledge
from task-specific experience, including
spatial grounding, contact control,
action coordination, and feedback-based
recovery. Its update is formulated as
\begin{equation}
    \mathcal{E}_{k+1}^{x}
    = \operatorname{Generalize}\left(
    \mathcal{E}_{k}^{x},
    \mathcal{E}_{k+1}^{s}(i)
    \right),
\end{equation}
where $\operatorname{Generalize}(\cdot)$
abstracts transferable knowledge from
task-specific experience.
This enables the Reflector to draw on
accumulated operational knowledge when
analyzing new but related manipulation tasks.

Together, these complementary components
constitute the long-term experiential memory
available for task $i$:
\begin{equation}
    \mathcal{E}_k
    =
    \left(
        \mathcal{E}_{k}^{s}(i),
        \mathcal{E}_{k}^{x}
    \right).
\end{equation}

\subsection{Evidence-based Experience Reflection}
\label{sec:reflection}

Robotic manipulation failures may arise from
different underlying causes, while similar
task outcomes often provide limited evidence
for identifying execution errors. Moreover,
unconstrained reflection may produce explanations
unsupported by observed interactions.
To address these challenges, we introduce
Evidence-based Experience Reflection, which
grounds failure diagnosis in concrete execution
evidence and accumulated experience.

After each execution attempt, the Reflector
analyzes the task instruction $g$, trajectory $\tau_k$, task outcome
$r_k$, current harness $\mathcal{H}_k$, and
experiential memory $\mathcal{E}_k$:
\begin{equation}
    d_k = \mathcal{R}_{\phi}
    (g,\tau_k,r_k,\mathcal{H}_k,\mathcal{E}_k),
\end{equation}
where $\mathcal{R}_{\phi}$ denotes the frozen
Reflector model and $d_k$ represents the
execution diagnosis.

Each reflection outcome is assigned to exactly one of the following five operational categories:
\begin{itemize}
    \item \textbf{Contact \& Pose Control:}
    Examines errors in physical interactions,
    including grasp configurations, contact
    positions, gripper operations, and
    object orientations.

    \item \textbf{Motion \& Coordination:}
    Focuses on improper action sequencing,
    arm coordination, and motion planning
    that may lead to unsuccessful or
    unsafe manipulation.

    \item \textbf{Task Protocol \& Completion:}
    Captures failures to satisfy task
    requirements, execution restrictions,
    success conditions, or termination
    criteria.

    \item \textbf{Spatial Grounding:}
    Addresses incorrect target identification,
    spatial relationship interpretation,
    and coordinate grounding that result
    in actions at unintended locations.

    \item \textbf{State Tracking \& Recovery:}
    Identifies deficiencies in tracking
    task progress, verifying action outcomes,
    interpreting feedback, and recovering
    from execution failures.
\end{itemize}


The Reflector identifies discrepancies between
intended and observed outcomes to produce
evidence-grounded diagnoses for subsequent
harness self-improvement.

\subsection{Recursive Harness Self-Improvement}
\label{sec:improvement}

Following evidence-based reflection, the framework
converts execution diagnosis into reusable
corrective guidance and refines the execution harness.

Given the diagnosis $d_k$ produced by the Reflector,
the framework identifies operations requiring
adjustment and specifies their execution conditions,
conditioned on the current harness and accumulated
experience:
\begin{equation}
    \Delta\mathcal{G}_k
    = \operatorname{Generate}
    (d_k,\mathcal{H}_k,\mathcal{E}_k),
\end{equation}
where $\Delta\mathcal{G}_k$ denotes candidate guidance
that translates identified execution issues into
actionable instructions for subsequent attempts.

The candidate guidance is integrated with existing
operational knowledge to update the harness:
\begin{equation}
    \mathcal{H}_{k+1}
    = \operatorname{Update}
    (\mathcal{H}_k,\Delta\mathcal{G}_k,\mathcal{E}_k).
\end{equation}
The update uses accumulated experience to retain
effective guidance, resolve inconsistencies, and
refine instructions based on execution evidence.
The refined harness $\mathcal{H}_{k+1}$ guides
subsequent Actor attempts, whose trajectories and
feedback support further reflection.
This cycle of execution, diagnosis, guidance
generation, and harness refinement enables
progressive improvement without updating model
parameters.

\definecolor{groupgray}{RGB}{242,242,242}
\definecolor{bestblue}{RGB}{225,237,250}

\begin{table}[t]
\centering
\centering
\caption{
Success rates (SR, \%) on RoboTwin 2.0 under
the Clean-to-Randomized (C2R) setting.
Methods are grouped into VLA models, WAMs,
and agentic frameworks.
Shots denotes the number of task-specific
demonstrations used.
Our method is highlighted in \colorbox{bestblue}{blue},
and the best result is boldfaced.
}

\label{tab:robotwin2_c2r}

\small
\setlength{\tabcolsep}{16pt}


\begin{tabular}{lcc}

\toprule
\textbf{Method} & \textbf{Shots} & \textbf{SR (\%)} \\
\midrule

\rowcolor{groupgray}
\multicolumn{3}{c}{\textit{Vision-Language-Action Models}} \\
\midrule
StarVLA~\cite{community2026starvlalegolikecodebasevisionlanguageaction}             & Full Set & 10.6 \\
Xiaomi-Robotics-0~\cite{cai2026xiaomirobotics0opensourcedvisionlanguageactionmodel} & Full Set & 18.2 \\
GR00T-N1.7~\cite{nvidia2025gr00tn1openfoundation}                                    & Full Set & 20.7 \\
X-VLA~\cite{zheng2026xvla}                                                          & Full Set & 20.9 \\
Spatial Forcing~\cite{li2026spatial}                                                & Full Set & 26.7 \\
$\pi_{0.5}$~\cite{intelligence2025pi05visionlanguageactionmodelopenworld}           & Full Set & 46.0 \\
LingBot-VLA~\cite{wu2026pragmaticvlafoundationmodel}                                & Full Set & 50.4 \\
GigaBrain-0.7~\cite{gigabrainteam2026gigabrain07scalingembodiedfoundation}          & Full Set & 67.9 \\
\midrule

\rowcolor{groupgray}
\multicolumn{3}{c}{\textit{World-Action Models}} \\
\midrule
FastWAM~\cite{yuan2026fastwamworldactionmodels}               & Full Set & 1.9  \\
X-WAM~\cite{guo2026unified4dworldaction}                      & Full Set & 25.8 \\
PatchWAM~\cite{wang2026actionworthpatchunified}               & Full Set & 66.7 \\
UniWAM~\cite{song2026uniwamunifiedworldactionmodel}           & Full Set & 68.3 \\
PICA-WAM~\cite{zou2026picawamvisualperturbationinvariance}    & Full Set & 74.7 \\
PrismWAM$^{*}$~\cite{xin2026prismwamdecouplingstaterepresentations} & Full Set & 75.5 \\
\midrule

\rowcolor{groupgray}
\multicolumn{3}{c}{\textit{Agentic Frameworks}} \\
\midrule
HarnessVLA (Codex)~\cite{zhang2026harnessvlasteeringfrozen}              & Full Set & 58.0 \\
HarnessVLA (CC)~\cite{zhang2026harnessvlasteeringfrozen}                 & Full Set & 58.4 \\
RoboDawn (GPT-6 Astra)~\cite{guo2026transferringintelligencevlmsrobotic} & 0        & 53.2 \\
RoboDawn (GPT-6 Astra)~\cite{guo2026transferringintelligencevlmsrobotic} & 1        & 73.6 \\
\rowcolor{bestblue}
\textbf{Ours} (GPT-6 Astra) & 0 & \textbf{82.4} \\
\bottomrule


\bottomrule
\end{tabular}
\end{table}
\section{Experiments}
\label{sec:experiments}

\subsection{Experimental Setup}
\noindent\textbf{Evaluation Benchmarks.}
We evaluate Embodied-Harness on three representative
simulation benchmarks: RoboTwin 2.0~\cite{pmlr-v306-chen26j}, RoboSPA~\cite{fan-etal-2026-robospa},
and RMBench~\cite{chen2026rmbenchmemorydependentroboticmanipulation}, alongside three real-world
manipulation tasks.
RoboTwin 2.0 includes 50 tasks evaluated
under the Clean-to-Randomized (C2R) setting.
RoboSPA comprises 24 fine-grained spatial
reasoning tasks and 32 long-horizon procedural
planning tasks, both evaluated at the highest
difficulty level (L5).
We additionally evaluate spatial reasoning
across L1--L4 to examine performance under
varying task difficulty.
RMBench contains 9 memory-dependent
manipulation tasks requiring information
retention across an episode.
Finally, three real-world tasks assess
the effectiveness of Embodied-Harness
in physical environments.

\noindent\textbf{Evaluation Protocol.}
We use GPT-6 Astra as the backbone model, with medium reasoning effort as the default setting. Each evaluation episode is limited to 45 interaction rounds.
For real-world experiments, we use the Piper robotic arm as the manipulation platform. We conduct 10 evaluation episodes per task on RoboTwin 2.0, RoboSPA, and real-world tasks, and 20 episodes per task on RMBench. We report success rates (SR) for each benchmark, and additionally report progress scores (PS) on RoboSPA long-horizon procedural planning tasks to measure the degree of partial task completion.

\subsection{Main Results}




\begin{table}[t]
\centering
\caption{
Success rates (SR, \%) on RoboSPA
fine-grained spatial reasoning at the
highest difficulty level (L5).
Avg. denotes the unweighted average
across five categories.
}

\label{tab:robospa_spatial}

\small
\setlength{\tabcolsep}{8pt}


\begin{tabular}{l*{6}{c}}
\toprule
\textbf{Method}
& GAC & SDE & CPI & RRR & CVR & \textbf{Avg.} \\
\midrule

RDT~\cite{liu2025rdtb}                                                    & 4.8  & 9.4  & 8.6  & 4.4  & 3.4  & 6.1  \\
GO-1~\cite{agibotworldcontributors2025agibotworldcolosseolargescale}      & 11.5 & 11.6 & 7.2  & 6.6  & 16.2 & 10.6 \\
$\pi_{0.5}$~\cite{intelligence2025pi05visionlanguageactionmodelopenworld} & 12.5 & 23.4 & 23.6 & 9.8  & 37.6 & 21.4 \\
X-VLA~\cite{zheng2026xvla}                                                & 25.3 & 34.6 & 8.8  & 14.2 & 36.6 & 23.9 \\

\midrule

RoboDawn~\cite{guo2026transferringintelligencevlmsrobotic}                & 80.0 & 68.0 & 74.0 & 82.0 & 24.0 & 65.6 \\

\rowcolor{bestblue}
\textbf{Ours}
& \textbf{100.0}
& \textbf{94.0}
& \textbf{92.0}
& \textbf{86.0}
& \textbf{98.0}
& \textbf{94.0} \\


\bottomrule
\end{tabular}
\end{table}


\begin{table*}[t]
\centering
\caption{
Success rates (SR, \%) and progress scores (PS, \%) on RoboSPA long-horizon procedural planning tasks at L5.
}

\label{tab:robospa_long}

\small
\setlength{\tabcolsep}{3pt}


\begin{tabularx}{\textwidth}{
    @{}l
    *{6}{>{\centering\arraybackslash}X}|
    *{6}{>{\centering\arraybackslash}X}@{}
}
\toprule

\multirow{2}{*}{\textbf{Method}}
& \multicolumn{6}{c|}{
    \cellcolor{groupgray}\textbf{SR (\%)}}
& \multicolumn{6}{c}{
    \cellcolor{groupgray}\textbf{PS (\%)}} \\

\cmidrule(lr){2-7}
\cmidrule(l){8-13}

& RPF & OFE & OCE & CAC & MIP & Avg.
& RPF & OFE & OCE & CAC & MIP & Avg. \\

\midrule

RDT~\cite{liu2025rdtb}
& 35.0 & 0.8 & 0.1 & 2.4 & 0.0 & 7.7
& 44.5 & 24.9 & 14.7 & 26.5 & 3.6 & 22.8 \\

GO-1~\cite{agibotworldcontributors2025agibotworldcolosseolargescale}
& 23.8 & 5.6 & 3.1 & 2.6 & 0.0 & 7.0
& 32.0 & 25.7 & 14.8 & 22.7 & 6.0 & 20.2 \\

$\pi_{0.5}$~\cite{intelligence2025pi05visionlanguageactionmodelopenworld}
& \textbf{65.0} & 26.1 & 12.0 & 12.5 & 0.0 & 23.1
& \textbf{77.3} & 62.1 & 28.0 & 48.0 & 11.4 & 45.4 \\

X-VLA~\cite{zheng2026xvla}
& 22.5 & 43.6 & 7.7 & 5.9 & 0.0 & 15.9
& 31.9 & 73.9 & 25.1 & 40.1 & 7.3 & 35.7 \\

RoboDawn~\cite{guo2026transferringintelligencevlmsrobotic}
& 47.5 & 60.0 & 48.6 & \textbf{35.0} & 16.0 & 41.4
& 53.0 & 75.0 & 63.3 & 63.5 & 27.2 & 56.4 \\

\rowcolor{bestblue}
\textbf{Ours}
& 57.5 & \textbf{66.2} & \textbf{58.6}
& \textbf{35.0} & \textbf{80.0} & \textbf{59.5}
& 60.5 & \textbf{81.0} & \textbf{72.6}
& \textbf{68.5} & \textbf{92.0} & \textbf{74.9} \\

\bottomrule
\end{tabularx}
\end{table*}

\begin{table*}[t]
    \centering
    \caption{
        Success rates (\%) on the nine memory-dependent
        manipulation tasks in RMBench.
    }
    \label{tab:rmbench}

    \small
    \setlength{\tabcolsep}{2pt}



\begin{tabular*}{\textwidth}{
    @{\extracolsep{\fill}}l*{7}{c}
    >{\columncolor{bestblue}}c@{}
}
        \toprule
        \multirow{2}{*}{\textbf{Task}}
        & \multicolumn{3}{c}{
            \cellcolor{groupgray}\textit{VLA Models}}
        & \multicolumn{3}{c}{
            \cellcolor{groupgray}\textit{WAMs}}
        & \multicolumn{2}{c}{
            \cellcolor{groupgray}\textit{Agentic Frameworks}} \\
        \cmidrule(lr){2-4}
        \cmidrule(lr){5-7}
        \cmidrule(l){8-9}


& \makecell{$\pi_{0.5}$\\
  \cite{intelligence2025pi05visionlanguageactionmodelopenworld}}
& \makecell{X-VLA\\
  \cite{zheng2026xvla}}
& \makecell{Mem-0\\
  \cite{chen2026rmbenchmemorydependentroboticmanipulation}}
& \makecell{LingBot-VA\\
  \cite{LiL-RSS-26}}
& \makecell{MemoryWAM\\
  \cite{yang2026memorywamefficientworldaction}}
& \makecell{MaP-WAM\\
  \cite{zhao2026memoryplansworldactionmodeling}}
& \makecell{RoboDawn\\
  \cite{guo2026transferringintelligencevlmsrobotic}}
& \textbf{Ours} \\

        \midrule

        Observe and Pick Up
        & 9 & 9 & 4
        & 13 & 27 & 19
        & \textbf{100} & \textbf{100} \\

        Rearrange Blocks
        & 13 & 13 & 89
        & \textbf{100} & \textbf{100} & 66
        & \textbf{100} & \textbf{100} \\

        Put Back Block
        & 11 & 18 & 90
        & \textbf{100} & \textbf{100} & \textbf{100}
        & 25 & \textbf{100} \\

        Swap Blocks
        & 24 & 16 & 67
        & 99 & \textbf{100} & 97
        & 0 & 90 \\

        Swap T
        & 15 & 3 & 14
        & 88 & 94 & \textbf{96}
        & 80 & 90 \\

        Battery Try
        & 16 & 26 & 28
        & 41 & 41 & 82
        & 35 & \textbf{100} \\

        Blocks Ranking Try
        & 6 & 1 & 18
        & \textbf{100} & \textbf{100} & 94
        & \textbf{100} & \textbf{100} \\

        Cover Blocks
        & 0 & 2 & 68
        & 79 & 98 & \textbf{100}
        & 60 & 85 \\

        Press Button
        & 0 & 0 & 0
        & 84 & 87 & \textbf{96}
        & 65 & 85 \\

        \midrule
        \textbf{Average}
        & 10.4 & 9.8 & 42.0
        & 78.2 & 83.0 & 83.3
        & 62.8 & \textbf{94.4} \\

        \bottomrule
\end{tabular*}
\end{table*}








\begin{table}[t]
    \centering
    \begin{minipage}[t]{0.46\textwidth}
        \vspace{0pt}
        \centering
        \caption{Success rates (\%) on three real-world robotic manipulation tasks.
Embodied-Harness outperforms $\pi_{0.5}$ on all three tasks, with an
average success rate of 96.7\% versus 40.0\%.}
        \label{tab:real_world}
        \small
        \setlength{\tabcolsep}{2pt}

 \begin{tabular}{l*{4}{c}}
        \toprule
        \textbf{Method}
        & \textbf{Arrangement}
        & \textbf{Placement}
        & \textbf{Stacking}
        & \textbf{Avg.} \\

        \midrule

        $\pi_{0.5}$
        & 40.0 & 50.0 & 30.0 & 40.0 \\

        \rowcolor{bestblue}
        \textbf{Ours}
        & \textbf{90.0}
        & \textbf{100.0}
        & \textbf{100.0}
        & \textbf{96.7} \\

        \bottomrule
\end{tabular}

    \end{minipage}\hfill
    \begin{minipage}[t]{0.51\textwidth}
        \vspace{0pt}
        \centering
        \caption{Ablation results on RoboSPA-Long (L5).}
        \label{tab:ablation}
        \small
        \setlength{\tabcolsep}{2pt}
        \begin{tabular}{lcc}
        \toprule
        \textbf{Method}
        & \textbf{SR (\%)}
        & \textbf{PS (\%)} \\
        \midrule
        \rowcolor{bestblue}
        \textbf{Embodied-Harness}
            & \textbf{59.5} & \textbf{74.9} \\
        \midrule

        w/o Reflector                     & 43.0 & 61.2 \\
        w/o Short-Term Working Memory     & 32.1 & 49.5 \\
        w/o Long-Term Experiential Memory & 48.5 & 65.6 \\
        w/o Evidence Grounding   & 52.3 & 67.9 \\

        \bottomrule
    \end{tabular}
    \end{minipage}
\end{table}
\noindent\textbf{RoboTwin 2.0 C2R.}
Table~\ref{tab:robotwin2_c2r} compares Embodied-Harness with representative vision-language-action models and world-action models, as well as agentic frameworks, on RoboTwin 2.0 C2R.
Our method achieves an overall success rate of \textbf{82.4\%}, outperforming all compared methods.
Under the same zero-shot setting and with the same GPT-6 Astra backbone, Embodied-Harness surpasses RoboDawn by 29.2 percentage points.
This comparison supports the effectiveness of our framework in improving manipulation performance beyond the underlying model's capabilities.

Our method also exceeds the strongest VLA and WAM baselines, GigaBrain-0.7 and PrismWAM, by 14.5 and 6.9 percentage points, respectively, and outperforms the best HarnessVLA variant by 24.0 percentage points.
These methods use the full demonstration set, whereas Embodied-Harness operates with zero demonstrations.
Together, these results demonstrate the effectiveness of Embodied-Harness on RoboTwin 2.0 C2R without task demonstrations or model parameter updates.

\begin{wrapfigure}{r}{0.48\textwidth}
  \vspace{-15pt}
    \centering
    \includegraphics[width=\linewidth]{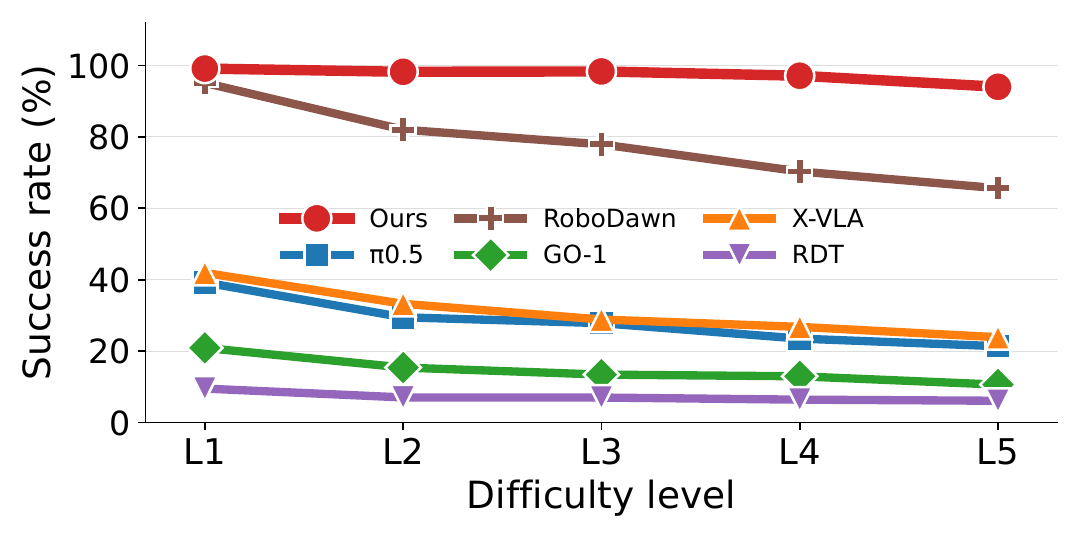}
    \caption{
         Success rates across five difficulty levels on RoboSPA fine-grained spatial reasoning tasks.
Embodied-Harness maintains 94--99\% success rates from L1 to L5,
whereas RoboDawn and VLA baselines generally decline with increasing difficulty.
    }
    \label{fig:robospa_difficulty}
\end{wrapfigure}
\noindent\textbf{RoboSPA-Spatial.}
Table~\ref{tab:robospa_spatial} presents
the results on fine-grained spatial reasoning
at the highest difficulty level (L5).
Embodied-Harness achieves an average SR of
94.0\%, outperforming the strongest baseline,
RoboDawn, by 28.4 percentage points.
Our method leads across all five categories,
with a particularly substantial improvement
on CVR from 24.0\% to 98.0\%.
Furthermore, Fig.~\ref{fig:robospa_difficulty}
shows consistently high performance across
all five difficulty levels, with only minor
degradation as task complexity increases.
In contrast, competing methods exhibit
more pronounced declines, highlighting
the robustness of Embodied-Harness in
challenging spatial reasoning scenarios.

\noindent\textbf{RoboSPA-Long.}
Long-horizon procedural planning requires agents
to track task progress and coordinate interdependent
operations over extended sequences.
As shown in Table~\ref{tab:robospa_long},
Embodied-Harness achieves the highest average
SR of 59.5\% and PS of 74.9\%,
surpassing RoboDawn by 18.1 and 18.5
percentage points, respectively.
Our method ranks first in four of five
categories for both metrics, including a tie
in CAC SR.
The largest gains occur in MIP, with improvements
of 64.0 and 64.8 percentage points in SR and PS
over the strongest baseline.
These results support the effectiveness of
Embodied-Harness in sustaining progress
and completing long-horizon manipulation tasks.

\noindent\textbf{RMBench.}
Table~\ref{tab:rmbench} reports results on nine
memory-dependent manipulation tasks.
Embodied-Harness achieves the highest average
success rate of 94.4\%, surpassing the strongest
baseline, MaP-WAM, by 11.1 percentage points.
It achieves 100\% success on five tasks and at
least 85\% on the remaining four, demonstrating
consistent performance across diverse scenarios.
These results support our hierarchical memory
design: Short-Term Working Memory preserves task
states and interaction histories for action
selection, while Long-Term Experiential Memory
retains task-specific and transferable knowledge
for reflection and iterative improvement.

\begin{wrapfigure}[14]{r}{0.48\textwidth}
    \vspace{-12pt}
    \centering
    \includegraphics[width=\linewidth]{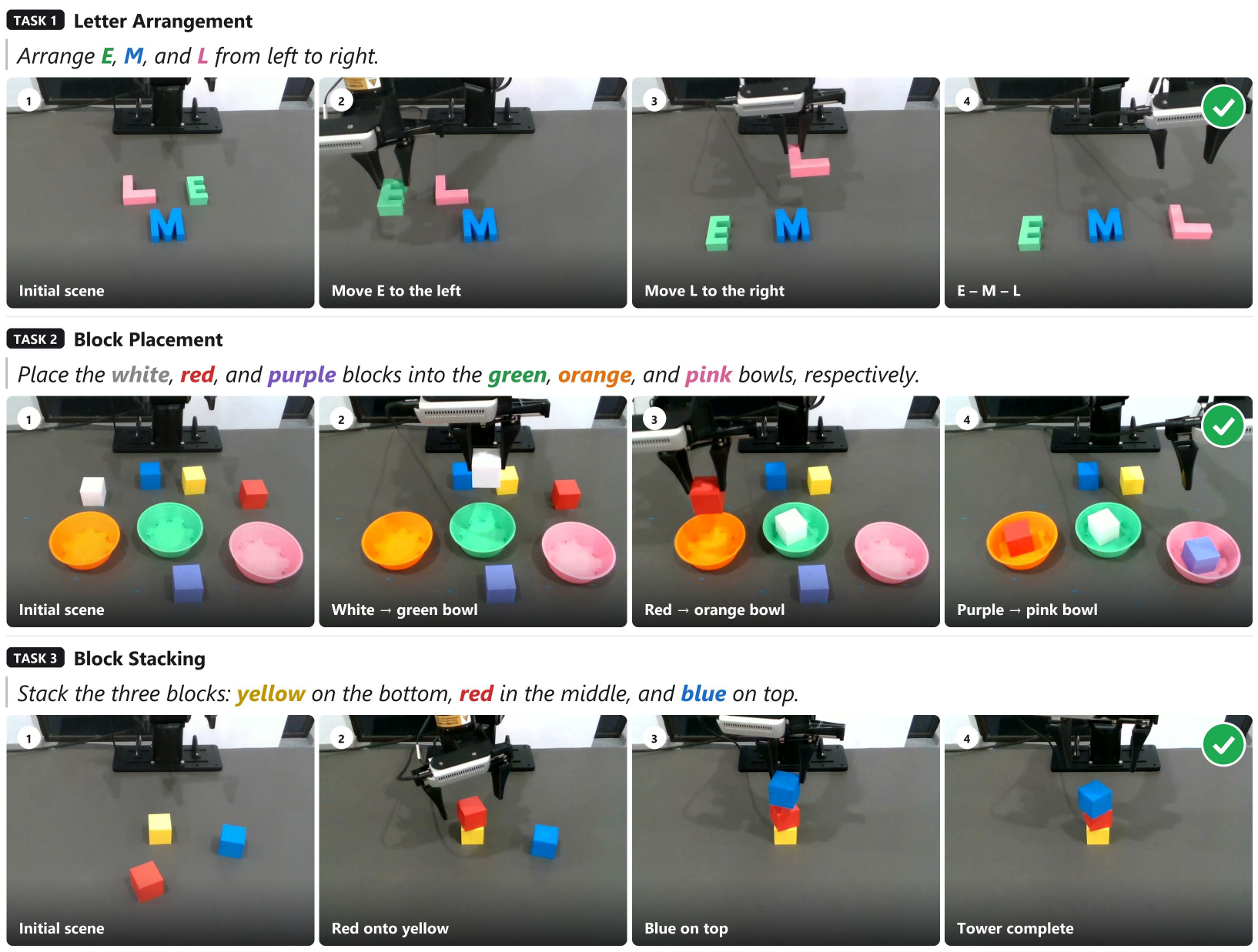}
    \caption{Execution examples on three real-world tasks.}
    \label{fig:real_world}
\end{wrapfigure}
\noindent\textbf{Real-World Experiments.}
We evaluate Embodied-Harness on three real-world
manipulation tasks requiring multi-step reasoning
and long-horizon execution (Fig.~\ref{fig:real_world}).
As reported in Table~\ref{tab:real_world},
Embodied-Harness achieves an average success rate
of 96.7\%, outperforming $\pi_{0.5}$ by 56.7
percentage points.
Our method achieves 90.0\% success on Letter
Arrangement and 100.0\% on both Block Placement
and Block Stacking, while $\pi_{0.5}$ achieves
40.0\% success on average.
These results demonstrate the effectiveness of
Embodied-Harness in real-world manipulation
requiring complex reasoning and coordinated
multi-step execution.





\begin{figure*}[t]
    \centering
    \begin{subfigure}[b]{0.49\textwidth}
        \centering
        \includegraphics[width=\linewidth]{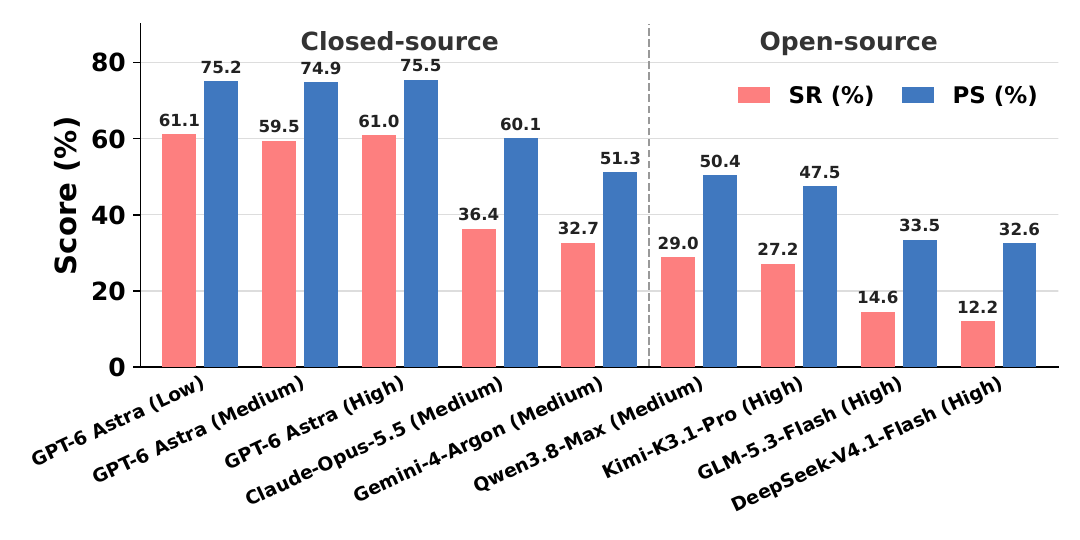}
        \caption{Backbone and reasoning effort.}
        \label{fig:model_ablation}
    \end{subfigure}\hfill
    \begin{subfigure}[b]{0.49\textwidth}
        \centering
        \includegraphics[width=\linewidth]{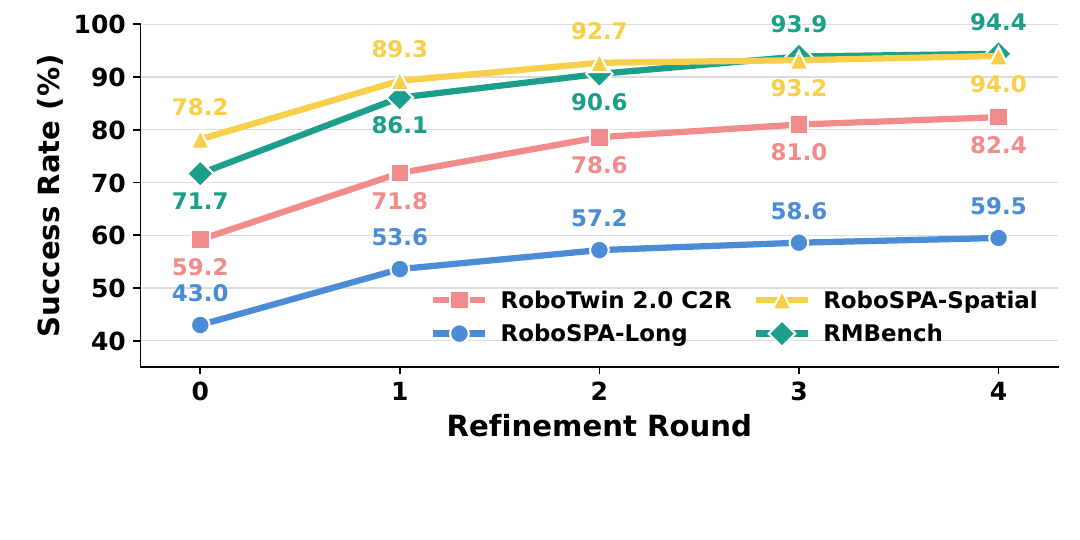}
        \caption{Recursive harness self-improvement.}
        \label{fig:refinement_curve}
    \end{subfigure}

    \caption{
\textbf{(a)} Effect of VLM backbone and reasoning effort
on RoboSPA-Long.
\textbf{(b)} Success rates across recursive harness
self-improvement rounds.
    }
    \label{fig:analysis}
\end{figure*}

\subsection{Ablation Studies}
\noindent\textbf{Effect of individual components.}
Table~\ref{tab:ablation} reports ablation results on RoboSPA-Long (L5), where all ablations degrade performance. Without Short-Term Working Memory, SR drops from 59.5\% to 32.1\% and PS from 74.9\% to 49.5\%, marking the largest decline and highlighting its role in maintaining execution context. Ablating the Reflector reduces SR to 43.0\%, while excluding Long-Term Experiential Memory yields 48.5\% SR, supporting the value of reflection and accumulated experience. Without Evidence Grounding, SR and PS decrease to 52.3\% and 67.9\%, respectively, underscoring the importance of execution evidence for reliable corrective guidance.

\noindent\textbf{Effect of VLM Backbone and Reasoning Effort.}
Figure~\ref{fig:model_ablation} compares the performance
of different VLM backbones and reasoning effort levels
on RoboSPA-Long.
GPT-6 Astra consistently achieves the strongest performance,
with SRs of 59.5--61.1\% and PSs of 74.9--75.5\%,
substantially outperforming other backbones.
Closed-source models generally outperform open-source
alternatives, with the strongest open-source model,
Qwen3.8-Max, achieving only 29.0\% SR and 50.4\% PS.
Notably, increasing the reasoning effort of GPT-6 Astra
from low to high yields no consistent improvement,
suggesting that backbone capability plays a more
critical role than reasoning effort.

\subsection{In-Depth Analysis}
\leavevmode 
\begin{wrapfigure}[23]{r}{0.48\textwidth}
    \vspace{-15pt}
    \centering
    \includegraphics[width=\linewidth]{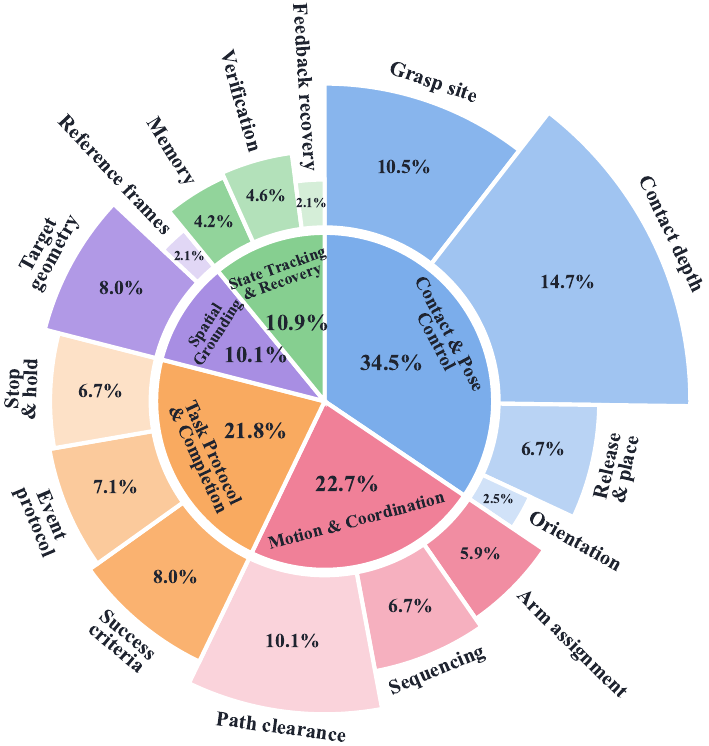}
    \caption{Distribution of reflection outcomes across operational categories.}
    \label{fig:improvement_taxonomy}
\end{wrapfigure}
\noindent\textbf{Recursive Harness Self-Improvement.}
Figure~\ref{fig:refinement_curve} shows
consistent performance gains across successive
harness refinement rounds.
From round 0 to round 4, success rates increase
from 59.2\% to 82.4\% on RoboTwin 2.0 C2R,
78.2\% to 94.0\% on RoboSPA-Spatial,
43.0\% to 59.5\% on RoboSPA-Long,
and 71.7\% to 94.4\% on RMBench.
Improvements are largest in the early rounds
and gradually diminish.
These results support the effectiveness of
repeated reflection and harness refinement
in improving manipulation performance without
updating model parameters.

\noindent\textbf{Distribution of reflection outcomes.}
Figure~\ref{fig:improvement_taxonomy} summarizes
the reflection outcomes across five categories.
Contact and pose control accounts for the largest
share (34.5\%), followed by motion and coordination
(22.7\%) and task protocol and completion (21.8\%).
Spatial grounding (10.1\%) and state tracking and
recovery (10.9\%) comprise the remaining outcomes.
This distribution suggests that reflection addresses
both low-level execution issues and higher-level
task requirements, providing corrective guidance
across multiple aspects of manipulation.

\section{Conclusion}
\label{sec:conclusion}
We presented Embodied-Harness, a dual-agent framework that converts execution experience into reusable operational guidance for robotic manipulation without updating model parameters. Hierarchical memory supports context-aware execution by the Actor and evidence-based reflection by the Reflector, enabling iterative harness improvement. Experiments on RoboTwin 2.0, RoboSPA, RMBench, and real-world tasks demonstrate its effectiveness, while ablations and iterative analyses substantiate the roles of memory, reflection, and evidence grounding. These findings establish harness refinement as a promising direction for improving embodied agents. Future work will investigate experience transfer across tasks and embodiments and more efficient execution and reflection.

\bibliographystyle{unsrtnat}
\bibliography{main}

\end{document}